\documentclass[10pt,twocolumn,letterpaper]{article}

\usepackage{iccv}
\usepackage{times}
\usepackage{epsfig}
\usepackage{amsmath}
\usepackage{amssymb}

\usepackage[bold]{hhtensor}
\usepackage{algpseudocode}
\usepackage{algorithmicx,algorithm}

\usepackage{booktabs}       
\usepackage{microtype}      
\usepackage[american]{babel}
\usepackage{graphicx}

\usepackage{multirow}
\usepackage{subcaption}

\newcommand{\squishlist}{
 \begin{list}{$\bullet$}
  { \setlength{\itemsep}{0pt}
     \setlength{\parsep}{1pt}
     \setlength{\topsep}{1pt}
     \setlength{\partopsep}{0pt}
     \setlength{\leftmargin}{1em}
     \setlength{\labelwidth}{1em}
     \setlength{\labelsep}{0.5em} } }

\newcommand{\squishend}{
  \end{list}  }

\def\n{\phantom{0}}
\usepackage{pifont}
\newcommand{\cmark}{\ding{51}}

\usepackage[breaklinks=true,bookmarks=false]{hyperref}

\iccvfinalcopy 

\def\iccvPaperID{****} 

\ificcvfinal\fi

\begin{document}

\title{Friends and Foes in Learning from Noisy Labels}

\author{Yifan Zhou\thanks{Equal contribution} , Yifan Ge\footnotemark[1] , Jianxin Wu \\
State Key Laboratory for Novel Software Technology \\
Nanjing University, Nanjing, China \\
{\tt\small zhouyifan@lamda.nju.edu.cn, lancelot1998kym@outlook.com, wujx2001@gmail.com}}

\maketitle
\ificcvfinal\thispagestyle{empty}\fi
\begin{abstract}
	Learning from examples with noisy labels has attracted increasing attention recently. But, this paper will show that the commonly used CIFAR-based datasets and the accuracy evaluation metric used in the literature are both inappropriate in this context. An alternative valid evaluation metric and new datasets are proposed in this paper to promote proper research and evaluation in this area. Then, friends and foes are identified from existing methods as technical components that are either beneficial or detrimental to deep learning from noisy labeled examples, respectively, and this paper improves and combines technical components from the friends category, including self-supervised learning, new warmup strategy, instance filtering and label correction. The resulting F\&F method significantly outperforms existing methods on the proposed nCIFAR datasets and the real-world Clothing1M dataset.
\end{abstract}

\section{Introduction}

Deep learning methods, although achieving excellent accuracy on a variety of vision tasks, are also notorious for their dependency on large training sets with precise annotations. The reality is that, however, this is a difficult-to-satisfy requirement. Instead, a small training set with precise annotations or a large training set with noisy labels are easy to obtain (\eg , through manual labeling or image search engines, respectively). Hence, learning from data with noisy labels has attracted increasing attention recently.

In spite of the many progresses made in recent years, the area of learning from noisy labels suffers from two fundamental drawbacks: \emph{improper dataset} and \emph{improper metric} for the evaluation of algorithms. In order to ablate various aspects of new algorithms, noisy CIFAR datasets~\cite{Tanaka18} were created by adding noise to the classic CIFAR~\cite{Krizhevsky09} datasets. A real-world data, Clothing1M~\cite{Xiao15}, was used to evaluate real-world performance. And, the accuracy was used in both settings as the performance metric.

Both drawbacks are connected to two improper assumptions implicitly assumed in most previous research: an \emph{ incorrect closed-world label noise} assumption and an \emph{incorrect balanced (clean) training set and test set} assumption.

One implicit assumption in existing algorithms is that the label noise is in the closed-world (or closed noise). For example, the noisy CIFAR datasets~\cite{Tanaka18} only contain closed noise; and DivideMix~\cite{LiJ20} treated examples that the model were not confident about as unlabeled ones, and subsequently used a semi-supervised learning method~\cite{Berthelot19}---which mostly assumed that these unlabeled examples come from \emph{only} these categories that we are interested in (\ie, in the closed-world). But this assumption does not fit the real-world. In the real-world Clothing1M~\cite{Xiao15} dataset, the `jacket' category contains lots of images of jeans (more than 50\%!), which is clearly a label noise in the open-world (or open noise, not in the 14 categories of Clothing1M). In fact, collecting images with noisy labels from search engines will \emph{inevitably contain open noise}, and it is natural to deduce that ignoring open noise will \emph{surely lead to inferior results}.

Another implicit previous assumption is that the training set (or its subset of those examples with clean labels) is balanced. The noisy CIFAR datasets~\cite{Tanaka18} are balanced (modulo slight sampling variances) in terms of those training examples with clean labels. But, because of the very existence of noisy labels, the clean labels will \emph{inevitably be imbalanced or even long-tailed}. The real-world Clothing1M~\cite{Xiao15} dataset even has an imbalanced test set! As is widely known in both vision and learning, accuracy is a poor metric when the test set is imbalanced or long-tailed, which will bias against under-represented categories with few examples and make evaluation results less trustworthy~\cite{Ouyang16}.

The drawbacks and assumptions are foes for developing, evaluating, \emph{comparing} and \emph{understanding} algorithms for learning from noisy labeled examples, and have implicitly affected most (if not all) previous efforts in this area. Hence, in this paper, we defend against these enemies by convening friends for learning from noisy labels:
\squishlist
	\item We create noisy CIFAR datasets that contain both open and closed noise, whose construction process is more scientific than that for existing ones;
	\item We show that mean average precision (mAP) is more suitable for evaluation, and advocate its adoption in this area;
	\item We show that self-supervised learning and warmup are effective in this task, but both require some special handling;
	\item We propose an improved PENCIL~\cite{Yi19}, where our instance filtering process is effective when noise exists and the revised PENCIL is good at correcting closed noisy labels.
\squishend

In short, by separating friends and foes, this paper contributes new datasets which are more suitable than existing ones, adopts and advocates a new evaluation metric as the replacement of the current accuracy metric, and proposes a method (called ``F\&F'') that achieves superior results than existing methods.

\section{Related Works}

First, we briefly review related works on learning from noisy labeled examples. Semi- and self-supervised learning are also discussed, as they are important tools in this area.

\subsection{Learning with noisy labels}

Most existing methods for training DNNs with noisy labels adopted at least one of the three approaches: to devise a loss function that is robust to noisy labels, to overcome overfitting via regularization, or to develop a process to relabel / correct the samples so that the data fed into a model can be less noisy.

PENCIL~\cite{Yi19} and SCE~\cite{Wang19} demonstrated that reversed KL-divergence, as the loss function, prevents a model from fitting corrupted labels by giving a small gradient to an example when the model and the label strongly disagree. \cite{Menon20} showed that a simple variant of gradient clipping can achieve the same goal. In contrast, MWNet~\cite{Shu19} fitted a mapping from loss value to sample weight used in the loss function instead of relying on a predefined scheme. Semi-supervised learning methods like DivideMix~\cite{LiJ20} can also be deemed a variant of reweighting.

AIR~\cite{Azadi16} designed a regularizer that encourages the model to learn from samples with similar representations to the established. \cite{Pereyra17} used negative entropy regularization to prevent the model from being too confident and overfitting the samples with corrupted labels. ELR~\cite{Liu20} proposed a simpler regularizer from an insight similar to AIR~\cite{Azadi16} and used it in the early-learning stage.

Previous works tried to model the noise with the help of a few clean samples to perform label correction~\cite{Vahdat17,LiY17, Lee18}. \cite{Tanaka18} proposed a framework to jointly optimize a model and pseudo-labels for each sample without access to any clean samples. On top of the framework, PENCIL~\cite{Yi19} adopted a robust loss and learned the pseudo-labels with stochastic gradient descent.

\subsection{Semi-supervised learning}

Semi-supervised learning deals with datasets where only a fraction of the samples are labeled. Most modern methods make use of the unlabeled data via regularization. Two kinds of regularization are the most commonly used: consistency-based regularization~\cite{Laine17} and entropy minimization~\cite{Grandvalet05}. MixMatch~\cite{Berthelot19} proposed a holistic framework by combining these two approaches with MixUp~\cite{Zhang18}.

\subsection{Self-supervised learning}

Self-supervised learning aims at learning a mapping from raw data to representation by exploiting the intrinsic structure of the data without resorting to labels. Various pretext tasks have been proposed for self-supervised learning on images, including Denoising Autoencoder, missing patches prediction~\cite{Vincent08}, solving Jigsaw puzzles~\cite{Noroozi16}, angle of rotation prediction~\cite{Gidaris18}, and more recently, determining if two augmented views come from the same original image~\cite{He20,Chen20}.

\section{F\&F: Datasets and Evaluation}

As aforementioned, existing noisy label learning datasets and the evaluation metric are both questionable. In this section, we will propose alternative ones for analyzing and validating algorithms.

\subsection{Proposing the nCIFAR datasets}

Two origins may produce images with noisy labels. First, when we use search engines and keywords to collect a dataset, like Clothing1M~\cite{Xiao15} and WebVision~\cite{Li17b}, both closed-world and open-world label noise exist. The datasets will also be imbalanced or long-tailed---the returned list of images for a common object will definitely be much longer than that for a rare object. Second, there are domains that are inherently difficult, where even labels manually provided by experts will be noisy, \eg, in medical imaging. In this scenario, open noise, closed noise and imbalance all naturally exist. Furthermore, label noise (both open and closed) often happens among categories that share certain degrees of similarity rather than occurring randomly.

We want the datasets we use to mimic real-world data, such that the proposed algorithms can be validly evaluated. However, the current noisy CIFAR datasets~\cite{Tanaka18} pose opposite properties: there are only closed-world noise (\eg, in noisy CIFAR10, all images come from the original 10 categories in CIFAR10); the subset of images with clean samples in the symmetric versions is balanced; label noise in the symmetric versions is uniform (\ie, regardless of category similarity), and label noise in the asymmetric versions for CIFAR100 is sequential in the category order (\ie, regardless of similarity and ad hoc). There is one work~\cite{Sachdeva21} trying to add open-world noise into CIFAR10, however, CIFAR10 is relatively simple and less effective in algorithm evaluation. Furthermore, \cite{Sachdeva21} did not take into account the category similarity and imbalance requirement. Hence, in order to evaluate algorithm effectiveness more reliably, we propose a principled and flexible approach to create new noisy CIFAR datasets (which we call nCIFAR) that satisfy all the above requirements.

To incorporate open-world noise, we use CIFAR100 as the source dataset to inject image into nCIFAR10, and Tiny ImageNet~\cite{Li17} (200 categories) to add open-world noise to nCIFAR100. Note that the categories in both pairs do not overlap. However, there are categories who share conceptual or visual similarity, which can either benefit or confuse the learning algorithms, depending on how they handle these noise images.

We can also find the semantic similarity between two categories by using word embeddings. For CIFAR categories, we extract the GloVe embedding vectors~\cite{Pennington14} (which are 50 dimensional) as their representation. For a Tiny ImageNet category / synset, we average the GloVe vectors of all the descriptive words in this synset. Because the word embedding encodes semantics, we can effectively use the cosine similarity between two categories' representation vectors to compute the similarity between them.

In our nCIFAR, a dataset is named in the convention ``nCIFAR10-$x$-$y$'' or ``nCIFAR100-$x$-$y$'' depending on whether CIFAR10 or CIFAR100 is the base, where $x$ and $y$ denote the percentage of open and closed noise rates, respectively. By specifying different $x$ and $y$, we can generate datasets with different noise levels flexibly. Suppose the base dataset has $n$ classes ($n=10$ or $n=100$) and the open noise source has $m$ classes ($m=100$ or $m=200$). We use a probabilistic sampling approach to generate them.

\squishlist
	\item First, we generate two similarity matrices. The closed-world similarity matrix $C$ is of size $n \times n$, and the open-world matrix $O$ is of $m \times n$. One entry $C_{ij}$ or $O_{ij}$ is computed using the cosine similarity.
	\item Second, since we often pick the top $k$ images returned by search engines in the real world, we generate $k$ ($k=5000$ for nCIFAR10 and $k=500$ for nCIFAR100, $nk=50000$) images in the generated dataset for every category. There will be $(1-x-y)nk$ clean examples in all categories, $xnk$ open-world noise examples, and $ynk$ closed-world noise images in total.
	\item To sample open noise images, we convert the similarity matrix $O$ by a softmax transformation with temperature $\tau_o$, then followed by a column-wise $\ell_1$ normalization, \ie, $$O'_{ij} = \frac{\exp(O_{ij}/\tau_o)}{\sum_{t=1}^m \exp(O_{tj}/\tau_o)} \,.$$ Because of the subsequent column-wise $\ell_1$ normalization processing, we do not need to complete the full softmax operation, and the exponential operation $\exp(O_{ij}/\tau_o)$ alone is enough. Then, the $j$-th column $O'_{:j}$ sums up to 1 and forms a valid distribution. Following this distribution, we sample $xk$ examples from the $m$ open noise source categories without replacement, and they form the open noise examples for the $j$-th category in the base dataset. We use $\tau_o=0.1$. Images are resized to $32 \times 32$.
	\item Similarly, to sample clean and closed noise samples, we transform $C$ to $C'$ by softmax with temperature $\tau_c$ and a subsequent $\ell_1$ normalization: $$C'_{ij} = \frac{\exp(C_{ij}/\tau_c)}{\sum_{t=1}^n \exp(C_{tj}/\tau_c)}\,.$$ Note that different $\tau_c$ leads to different proportion of sampled clean examples, and the relationship between them (value of $\tau_c$ and the number of clean examples) is monotonic, which is easily proved. Hence, we can use a binary search to find the $\tau_c$ value that leads to $(1-x-y)nk$ clean samples in total. Because the number of closed-world examples (clean plus closed noise) is $(1-x)nk$, the binary search terminates when the \emph{average} of the diagonal entries of $C'$ sum up to $\frac{1-x-y}{1-x}$. Also note that the binary search process has only trivial computational cost. Then, we use $C'_{:j}$ to sample $(1-x)k$ examples from all $n$ base categories to the $j$-th category according to $C'_{:j}$ with the proper $\tau_c$, which include \emph{both clean and closed noise samples}.
	\item Both training sets (nCIFAR10 and nCIFAR100) have size $nk=50000$. The test sets are the original CIFAR10 / 100 test sets, which are balanced.
\squishend

Note that the training datasets generated in our approach are imbalanced (different categories have different number of clean and closed noise samples), and categories with higher similarity scores are more likely to generate noisy samples. These properties make our nCIFAR datasets suitable in learning with noisy labeled examples.\footnote{It is desirable to have different numbers of open noise in different categories. However, since we have no clue on how open set noise distributes in general, we sample $xk$ open noise in every category.}

\subsection{Advocating the mAP evaluation metric}

\begin{figure}
	\centering
	\includegraphics[width=\columnwidth]{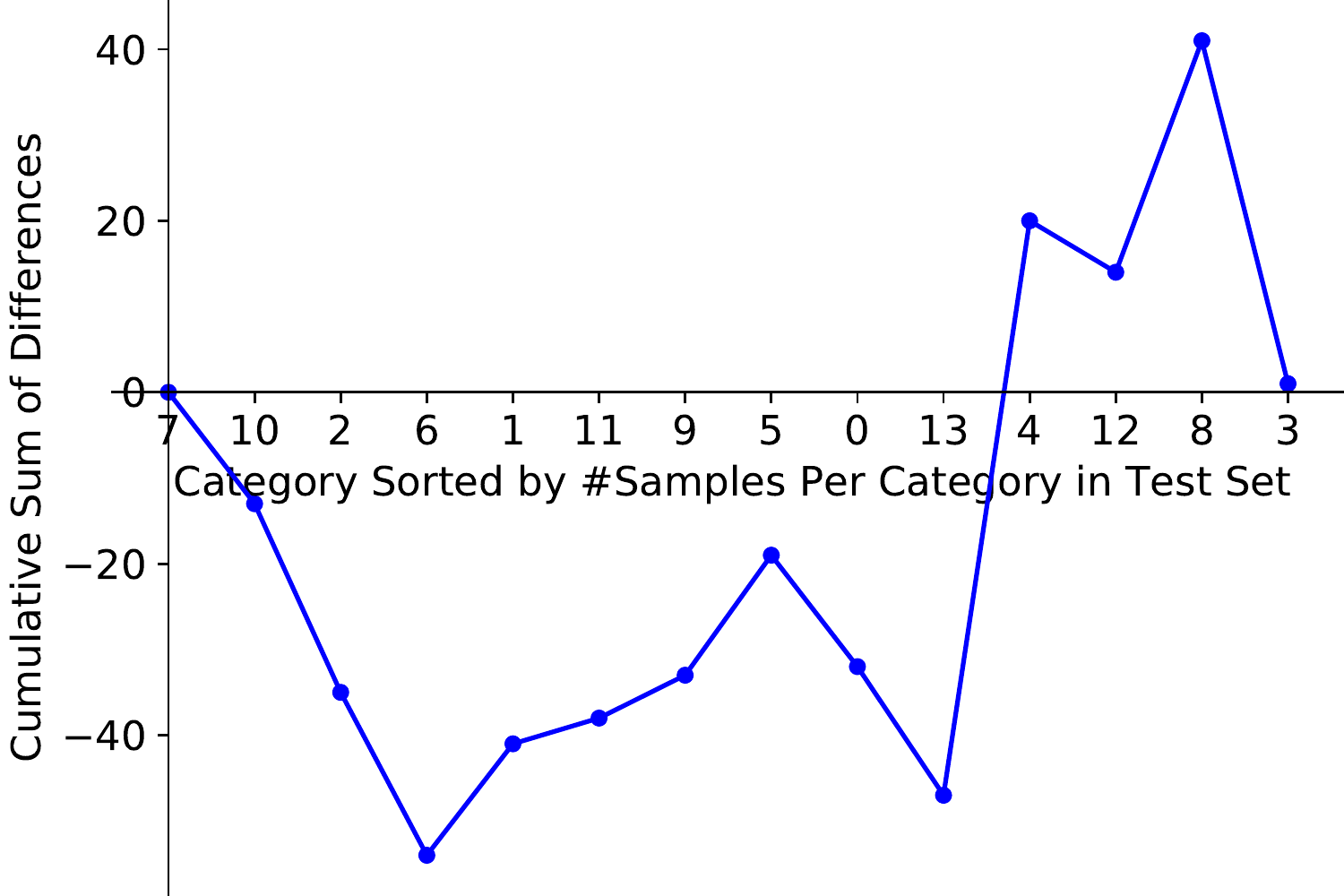}
	\caption{Comparing two classifiers in the imbalanced Clothing1M dataset across different categories.}
	\label{fig:mAP}
\end{figure}

Although imbalance is critical in making a good dataset, it causes difficulty in both training and evaluation. Suppose in an imbalanced binary recognition problem, the ratio between number of examples in two categories A and B is 99:1 in both training and test sets. Then, when accuracy is used as the evaluation metric, a model tends to predict all examples as A, and achieving 99\% accuracy. The minority class, B, is entirely sacrificed---hence, high accuracy does not necessarily mean good model in an imbalanced task.

A real-world dataset with label noise is imbalanced in the training set, and very likely also imbalanced in the test set, such as in Clothing1M~\cite{Xiao15}. Fig.~\ref{fig:mAP} compares two models (\texttt{Aug} for AugDesc~\cite{Nishi21} and \texttt{F\&F} for the proposed method) on Clothing1M (row 6 and 8 in Table~\ref{table: clothing1m}, respectively). They have almost identical accuracy (74.93\% for \texttt{Aug} and 74.92\% for \texttt{F\&F}). Although we do not know the number of \emph{clean} training examples in each category, it is reasonable to rank the categories by their size in the test set in the ascending order. Let $d_i$ denote the number of correctly predicted examples of \texttt{Aug} minus that of \texttt{F\&F} in the $i$-th ranked category, and $s_i$ the cumulative sum of $d_i$ (\ie, $s_i=\sum_{j=1}^i d_j$).

Fig.~\ref{fig:mAP} plots $s_i$, and an obvious trend shown in Fig.~\ref{fig:mAP} is that the curve is below the zero line in the left half (categories with fewer examples) and above it in the right half (categories with more examples). That is, \texttt{Aug} sacrificed accuracy of categories with fewer (clean) examples to boost the accuracy of other categories with more examples. Again, a higher accuracy value does \emph{not} necessarily mean a better model in an imbalanced dataset.

Following practices in evaluating imbalanced datasets, we advocate that we must use mean average precision (mAP) to evaluate models for noisy label learning, especially when the test set is imbalanced. mAP is not affected by the imbalance property~\cite{Ouyang16}. In this example, \texttt{F\&F} has an mAP of 75.22\%, outperforming \texttt{Aug} (73.25\%) by a large margin. As shown in Fig.~\ref{fig:mAP}, \texttt{F\&F} indeed acted more robustly than \texttt{Aug} across different categories.

\section{F\&F: Techniques}

Unlike earlier methods that often focus on one technical aspect (\eg, loss function), recent methods often build systems that integrate multiple techniques. For example, DivideMix~\cite{LiJ20} integrates techniques like semi-supervised learning, co-training, etc. Indeed, learning from noisy labels is a complex task, and has \emph{a strong systems flavor}. In this section, we propose a few techniques and build an F\&F system, which is based on DivideMix~\cite{LiJ20} and PENCIL~\cite{Yi19}.

\subsection{Self-supervised backbones \& freezed warmup}

\begin{table}
	\centering
	\begin{tabular}{c|c|l|c}
		\hline
		                                & backbone                & epoch     & accuracy         \\ \hline
		\multirow{8}{*}{warm-up policy} & \multirow{3}{*}{random} & \n0 + \n5 & 66.22            \\
		                                &                         & \n0 + 30  & 67.94            \\
		                                &                         & \n0 + 35  & 67.72            \\ \cline{2-4}
		                                & \multirow{2}{*}{ssl-1b} & \n0 + \n5 & 67.42            \\
		                                &                         & 30 + \n5  & 68.22            \\ \cline{2-4}
		                                & \multirow{3}{*}{ssl-2b} & \n0 + \n5 & 68.34            \\
		                                &                         & 30 + \n0  & 67.59            \\
		                                &                         & 30 + \n5  & $\textbf{68.38}$ \\ \hline
	\end{tabular}
	\caption{DivideMix on nCIFAR100-0.3-0.3. `random': randomly initialized backbones, `ssl-1b':  one self-supervised backbone (ssl) in both branches, `ssl-2b': two different ssl backbones. `$u$+$v$': $u$ freezed plus $v$ normal warmup epochs.}
	\label{table: warm-up policy}
\end{table}

Self-supervised learning has been shown to learn good representation without labels, a property that is conceptually particularly useful in our task (by throwing away all labels). Our first improvement (\emph{friend}) is to use self-supervised learning algorithms (SimCLR~\cite{Chen20} for nCIFAR and MoCo~\cite{He20} for Clothing1M) to learn initial backbone networks.

Although self-supervised learning has been applied to this domain~\cite{Mandal2020}, we argue that in complex methods like DivideMix which have 2 (or more) branches, \emph{diversity} is in fact essential---we have to use two \emph{different} self-supervised backbones to initialize them.

Another improvement (\emph{friend}) is the warmup of backbones. Before starting to train the models, a few warmup epochs are often needed, which updates all parameters in DivideMix. We argue  that it is prone to overfitting, and propose to first warmup $u$ epochs \emph{by freezing the backbone}, and then $v$ normal epochs that updates all parameters.

As the ablative experiments on nCIFAR100-0.3-0.3 shown in Table~\ref{table: warm-up policy}, both self-supervised learning (our diverse one in particular) and freezed warmup are friends for learning with noisy labels.\footnote{The nCIFAR test set is balanced, hence either accuracy or mAP can be used here.}

Self-supervised backbones, freezed and normal warmup, plus \emph{a smaller number} (compared to the original DivideMix) of epochs of DivideMix training form the stage 1 of F\&F.

\subsection{Improved PENCIL \& instance filtering}

The first stage's output model initializes that of the second stage in F\&F. In the first stage, DivideMix treats instances with large loss (\ie, less confident ones) as unlabeled examples and adopts semi-supervised learning on them. However, label correction learns or relabels some closed-world noise instances, and can significantly improve the accuracy~\cite{Yi19,Zheng21,Mandal2020}. Hence, label correction is another \emph{friend}.

We adopt the PENCIL framework~\cite{Yi19} in stage 2. However, it is difficult to tune the hyperparameters in PENCIL and it is rather complex. We simplify it and improve it.

Consider a $c$-class classification problem. One example $\vec{x}$ has a hard label $\hat{\vec{y}}$ in the one-hot encoding: $\vec{1}^T\hat{\vec{y}}=1$ and $\hat{\vec{y}} \in {\{0, 1\}}^{c}$, where $\vec{1}$ is a vector of all 1's. Note that the label is noisy, and hence can be wrong.

In PENCIL, a soft label $\vec{y}^d$ is defined, where $y^d_i$ is the estimated probability of $\vec{x}$ in the $i$-th class. That is, $\vec{y}^d$ forms a valid distribution. Note that $\vec{y}^d$ is not constant, but \emph{updated} by SGD (\ie, being corrected). For $\vec{x}$, the model predicts (after softmax) another distribution $f(\vec{x})$, and the Kullback-Leibler divergence $\mathrm{KL}(f(\vec{x}),\vec{y}^d)=\sum_{i=1}^c f_i(\vec{x})\log (f_i(\vec{x})/y^d_i)$ is PENCIL's first loss term. 

Its second loss term measures the compatibility between hard and soft labels, as $\mathrm{CE}(\hat{\vec{y}},\vec{y}^d)=-\sum_{i=1}^c \hat{y}_i\log y_i^d$. 

The third PENCIL loss term is the entropy of the prediction, $H(f(\vec{x}))=-\sum_{i=1}^c f_i(\vec{x})\log f_i(\vec{x})$.

Hence, the overall PENCIL loss is
\begin{equation}
	\frac{1}{c}\mathrm{KL}(f(\vec{x}),\vec{y}^d) + \alpha \mathrm{CE}(\hat{\vec{y}},\vec{y}^d) + \frac{\beta}{c}H(f(\vec{x})) \,,
\end{equation}
where $\alpha$ and $\beta$ are hyperparameters.

PENCIL is sensitive to the hyperparameters: $\alpha$, $\beta$, and in particular the learning rate to update $\vec{y}^d$. When the learning rate for the backbone is 0.01, the learning rate for updating $\vec{y}^d$ \emph{varies in different datasets} (which can be as large as 10000), and it is difficult to find a proper value for it. When these hyperparameter values are not suitable, the PENCIL training process often breaks down.

In our F\&F, we first \emph{remove} the compatibility (second) loss term (\ie, $\alpha=0$), and \emph{fix} $\beta=1$. Then, after removing these two hyperparameters, the loss is greatly simplified after simple algebraic manipulation: It becomes a reverse cross-entropy loss:
\begin{equation}
	\mathrm{CE}(f(\vec{x}),\vec{y}^d)= - \sum_{i=1}^c f_i(\vec{x}) \log y^d_i \,.
\end{equation}

In the original PENCIL, the gradient for the soft labels ($\vec{y}^d$) involves $1/y_i^d$, which is very unstable. The gradient in F\&F becomes simple: $\vec{y}^d - f(\vec{x})$, which greatly \emph{stabilizes} its updating, and we do \emph{not} need to tune the learning rate for updating $\vec{y}^d$.

The final friend that F\&F counts on is instance filtering, which has been shown to be effective~\cite{Sharma20,Pleiss20,Wu20,Mandal2020} in the presence of open noise. In PENCIL (and our improved version), closed noisy instances are in fact useful when the model is confident on them. Hence, we believe that \emph{removing low-confidence examples} in \emph{both} open and closed noise will be helpful.

When the prediction $f(\vec{x})$ has a large entropy, its confidence is low, and vice versa. In stage 2, we compute the prediction's entropy of all training examples, and fit a Gaussian mixture model (GMM) with 2 components---the first component is considered as high- and the second low-confidence. We can easily compute the posterior probability of an example belonging to the confident mode, and keep this example if the probability is larger than a threshold $p_e$. The rest examples are \emph{not} used. Note that this filtering is similar to that in DivideMix. However, examples with low-confidence are treated as unlabeled and still used in DivideMix, while we ignore them in F\&F. This instance filtering process is repeated for each of the 2 branches independently \emph{in every epoch}. That is, different subsets are filtered per epoch.

Note that when $p_e=0$, there is no instance filtering at all. We ablated on nCIFAR100-0.3-0.3. From Table~\ref{table: ablation_study_and_confident_samples}, it is clear that removing examples that are not confident ($p_e=0.5$) always outperforms using all examples ($p_e=0$). That is, instance filtering is an effective approach.

\begin{table}
	\centering
	\begin{tabular}{c c c c |c}
		\hline
		clean  & open   & closed & $p_e$         & accuracy         \\ \hline\hline
		\cmark & \cmark & \cmark & \phantom{0.}0 & 69.04 $\pm$ 0.25 \\
		\cmark & \cmark & \cmark & 0.5           & 69.69 $\pm$ 0.22 \\ \hline
		\cmark & \cmark &        & \phantom{0.}0 & 69.31 $\pm$ 0.24 \\
		\cmark & \cmark &        & 0.5           & 69.80 $\pm$ 0.28 \\ \hline
		\cmark &        & \cmark & \phantom{0.}0 & 69.35 $\pm$ 0.27 \\
		\cmark &        & \cmark & 0.5           & 69.90 $\pm$ 0.08 \\
		\hline\hline
		\cmark &        &        & n/a           & 71.65 $\pm$ 0.06 \\
		\cmark & \cmark & \cmark & n/a           & 66.61 $\pm$ 0.26 \\
		\hline
	\end{tabular}
	\caption{Effects of different types of noise and instance filtering on nCIFAR100-0.3-0.3. 5-time average reported. In the last two rows, models were trained with cross entropy without considering label noise, hence $p_e$ is not applicable.}
	\label{table: ablation_study_and_confident_samples}
\end{table}

Because we have groundtruth clean labels in nCIFAR for the entire dataset, we can manually remove either open noise or closed noise from the training set in stage 2 of F\&F for ablation. When open noise exists, instance filtering improves roughly 0.5--0.6\% in the first two groups in Table~\ref{table: ablation_study_and_confident_samples} and the standard deviations are all $>0.2$. In the third group, closed noise exists but open noise does not, where instance filtering improves 0.65\% and the standard deviation is much smaller (only 0.08), \ie, more stable. And, the sixth row has the highest accuracy, where open noise has been removed and instances have been filtered.

In the last group, we also trained networks using only the classic cross entropy loss (\ie, without noise handling). When only the clean samples are used, its accuracy is 71.65\%. When all examples are used (including both types of noisy examples), its accuracy is 66.61\%.

\begin{algorithm*}
	\caption{The proposed F\&F framework}
	\label{algorithm: FF_algorithm}
	\begin{algorithmic}[1]
		\State \Comment Stage 1
		\State Learn two backbone networks using self-supervised learning, and use them to initialize the two branches
		\State For each backbone, perform freezed and normal warmup
		\State Run DivideMix for a few epochs
		\State \Comment Stage 2
		\For {$t = 1$ to the number of training epochs}
			\For {each of the two branches}
				\State Compute entropy values for all training examples
				\State Fit a GMM with 2 components to model the entropy distribution
				\State Compute the posterior probability for all training examples
				\State Filter instances using the posterior probability, where $p_e$ is the threshold
				\State Run improved PENCIL on \emph{only} examples not filtered away \emph{in this epoch}
			\EndFor
		\EndFor
	\end{algorithmic}
\end{algorithm*}

These observations indicate that
\squishlist
	\item Closed noise is more friendly than open noise, because many closed noisy labels will be corrected by our improved PENCIL (\ie, becoming clean).
	\item When we use all samples (like in DivideMix), \ie, without instance filtering ($p_e=0$), both closed and open noise will be harmful.
	\item When we only keep confident samples ($p_e=0.5$), even open noise will not significantly impoverish the model.
	\item Both instance filtering and label correction are effective. They bring close to 1 percentage point improvement when combined in F\&F (69.04\% vs. 69.90\%).
	\item F\&F has significantly bridged the accuracy gap caused by noisy labels: 66.64\% when without noise handling, 69.90\% for F\&F, and 71.65\% for clean data.
\squishend

Finally, the two branches in stage 2 are \emph{independent} of each other. We are in fact training two different networks separately and then average their predictions.

\subsection{The F\&F framework}

Pseudo-code for the overall F\&F framework is shown in Algorithm~\ref{algorithm: FF_algorithm}. For more details, we will publish the code of F\&F.

It is also worth noting that the second stage runs much faster than the first stage. We recorded the timing statistics on the nCIFAR100-0.2-0.2 dataset, in which the first stage took 9.53 hours, while stage 2 only took 0.27 hours.

That is, the combination of instance filtering and label correction in stage 2, although only used 2.9\% running time of that of stage 1, can lead to considerable improvements in accuracy.

\section{Experimental Settings and Results}\label{sec:results}

In this section, we first describe the settings in our experiments, then present experimental results on both the proposed nCIFAR datasets and the real-world Clothing1M dataset. Although the noisy CIFAR datasets from~\cite{Tanaka18} are inappropriate, we also report F\&F's results on them, in order to facilitate comparisons with existing methods. We will publish the code to generate the proposed nCIFAR datasets, and the exact versions used in our experiments will be published, too. We will also publish code of the F\&F system.

\subsection{Detailed data and algorithmic settings}

\textbf{Datasets.} In this paper, we use 5 versions of the proposed nCIFAR datasets: nCIFAR10-0.1-0.1, nCIFAR10-0.2-0.2, nCIFAR100-0.1-0.1, nCIFAR100-0.2-0.2 and nCIFAR100-0.3-0.3. The last dataset has only 40\% clean samples. nCIFAR10 are relatively easy, which are suitable for sanity-check of algorithms. nCIFAR100 are more difficult, and are useful for evaluating, comparing, analyzing and understanding algorithms. Note that although we set the percentage of open noise the same as that of the closed noise, it is trivial to generate nCIFAR datasets with different ratios for them.

\textbf{Baseline methods.} In terms of baseline methods, we mainly compare with 2 existing methods: DivideMix~\cite{LiJ20} and AugDesc~\cite{Nishi21}. DivideMix is a strong method and many recent advances in learning from noisy labels are developed based on it. AugDesc is the current top performer.\footnote{https://paperswithcode.com/sota/image-classification-on-clothing1m}

For both methods, we used the codes published by their respective authors. We use their hyperparameter settings without modifications.

\textbf{F\&F: shared.} We update network parameters and soft labels using SGD with a momentum of 0.9. We set weight decay of network parameters to $5 \times 10^{-4}$ for CIFAR based datasets and $1 \times 10^{-3}$ for Clothing1M. We do not apply weight decay for updating soft labels.

\textbf{F\&F stage 2: Improved PENCIL.} As aforementioned, the hyperparameter setting in F\&F's stage 2 is much simpler than that in the original PENCIL~\cite{Yi19}. We used the \emph{same} setting in \emph{all} our experiments in this part. In the improved PENCIL training, we \emph{always} used 0.5 as the initial learning rate to update soft labels $\vec{y}^d$. For other network parameters, the learning rate was $2 \times 10^{-4}$ for all CIFAR based datasets and $1 \times 10^{-5}$ for Clothing1M. Improved PENCIL always ran 20 epochs, and both learning rates were reduced by a factor of 10 midway (\ie, after 10 epochs).

\textbf{F\&F stage 2: Instance filtering.} Since we have shown the effectiveness of instance filtering in ablative studies (cf. Table~\ref{table: ablation_study_and_confident_samples}), we always used $p_e=0.5$ in instance filtering. For all CIFAR based datasets, label correction with instance filtering was conducted on all the training samples. For Clothing1M, we sampled a subset containing about 260k images to apply label correction, following the original PENCIL~\cite{Yi19}. Note that this subset is balanced when considering the noisy labels. However, if we have access to clean labels, this subset will \emph{not} be balanced.

\textbf{F\&F stage 1: Self-supervised learning.} For noisy CIFAR from~\cite{Tanaka18}, we used SimCLR~\cite{Chen20} to train PreActResNet-18~\cite{He16b} on CIFAR for 1000 epochs. Because these datasets share the \emph{same} unlabeled images for CIFAR10 / 100 regardless of the noise type or noise ratio, we only need to train a pair of backbones with the original CIFAR10 / 100 datasets, and they can be shared among experiments on different noisy CIFAR datasets.

On our nCIFAR datasets, SimCLR (1000 epochs) was still used. However, because different nCIFAR10 or nCIFAR100 datasets have different sets of unlabeled training images, it is essential to train a pair of backbones for each nCIFAR dataset separately.

On Clothing1M, we used MoCo~\cite{He20} to train two different backbones of ResNet-50~\cite{He16a}, initialized with the official pretrained weights trained on ImageNet (800 epochs), then trained on Clothing1M for 200 epochs.

\textbf{F\&F stage 1: Warmup.} For noisy CIFAR in~\cite{Tanaka18}, in warmup we trained the classifiers for 30 epochs with the backbones frozen. The learning rate was $5 \times 10^{-2}$, and the batch size was 128. Because we only experimented with very high noise rate here, there was no normal warmup training in order to avoid overfitting.

For nCIFAR, we first train the classifiers for 30 epochs with the backbones frozen and then warm-up the whole network for 5 epochs. The learning rate was $5 \times 10^{-2}$, and the batch size was 128.

On Clothing1M, we sampled a (noisy) label-balanced subset containing about 260k images, and trained the classifiers for 10 epochs with the backbones frozen. We used a learning rate of $4 \times 10^{-3}$, and a batch size of 128.

\textbf{F\&F stage 1: DivideMix.} On all datasets, the temperature of label sharpening and the threshold of small-loss sample filtering were both fixed to 0.5 in all experiments.

For noisy CIFAR in~\cite{Tanaka18}, the DivideMix training part in stage 1 used the same parameters as those in the original DivideMix.

For nCIFAR, the learning rate and batch size were the same as in warmup. $\lambda_u$, strength of the consistency loss in DivideMix was set to 100 for nCIFAR100, and 0 for nCIFAR10, following the original DivideMix. $\alpha$ in mixup was set to 0.5. The networks were trained for 200 epochs with the learning rate reduced by a factor of 10 after half of the epochs.

On Clothing1M, we set $\alpha$ to 0.2, and $\lambda_u$ to 0 (which follows the original DivideMix). We trained the networks for 30 epochs. In each epoch we sampled 2000 (noisy) label-balanced mini-batches, and the batch  size was 64. The learning rate was set to $2 \times 10^{-3}$, and was reduced by a factor of 10 after 15 epochs.

For all these methods, we first sum the logits of the two branches, then apply a softmax transformation to predict the final results. When standard deviations are reported, we ran the experiments 5 times and report the mean accuracy or mAP, plus its standard deviation.

\begin{table}
	\centering
	\begin{tabular}{cc|cc}
		\hline
		Selection criterion            & Method      & mAP              & accuracy         \\ \hline
		\multirow{4}{*}{last}          & DivideMix   & 72.27            & 74.03            \\
		                               & AugDesc     & 72.92            & $\textbf{75.18}$ \\
		                               & F\&F stage1 & 72.39            & 73.83            \\
		                               & F\&F stage2 & $\textbf{75.31}$ & 74.89            \\ \hline
		\multirow{4}{*}{best accuracy} & DivideMix   & 72.22            & 74.19            \\
		                               & AugDesc     & 73.25            & $\textbf{74.93}$ \\
		                               & F\&F stage1 & 73.75            & 74.54            \\
		                               & F\&F stage2 & $\textbf{75.22}$ & 74.92            \\ \hline
		\multirow{4}{*}{best mAP}      & DivideMix   & 73.15            & 74.33            \\
		                               & AugDesc     & 73.42            & 74.87            \\
		                               & F\&F stage1 & 74.78            & 75.00            \\
		                               & F\&F stage2 & $\textbf{75.42}$ & $\textbf{75.10}$ \\ \hline
	\end{tabular}
	\caption{Comparison with state-of-the-art methods on Clothing1M. We ran author-provided codes to generate DivideMix and AugDesc results.}
	\label{table: clothing1m}
\end{table}

\begin{table*}
	\centering
	\setlength{\tabcolsep}{2.5pt}
	\begin{tabular}{c|cc|ccc}
		\hline
		Dataset & nCIFAR10-0.1-0.1                       & nCIFAR10-0.2-0.2                       & nCIFAR100-0.1-0.1                     & nCIFAR100-0.2-0.2                     & nCIFAR100-0.3-0.3                     \\ \hline
		DivideMix   & 93.88 $\pm$ 0.22              & 92.00 $\pm$ 0.32              & 75.76 $\pm$ 0.21          & 73.55 $\pm$ 0.49          & 67.66 $\pm$ 0.32          \\ \hline
		F\&F stage1 & 94.16 $\pm$ 0.07              & 92.15 $\pm$ 0.34              & 76.04 $\pm$ 0.21          & 73.68 $\pm$ 0.28          & 68.29 $\pm$ 0.12          \\ \hline
		F\&F stage2 & \textbf{94.73 $\pm$ 0.02}     & \textbf{92.84 $\pm$ 0.26}     & \textbf{77.44 $\pm$ 0.18} & \textbf{75.29 $\pm$ 0.21} & \textbf{69.69 $\pm$ 0.22} \\ \hline
	\end{tabular}
	\caption{Comparison between F\&F and DivideMix in terms of accuracy (\%) on the proposed nCIFAR datasets.}
	\label{table:nCIFAR}
\end{table*}

\begin{table*}
	\centering
	\begin{tabular}{cc|cc|cc}
		\hline
		                                         &                                       & \multicolumn{2}{c|}{CIFAR10} & \multicolumn{2}{c}{CIFAR100}                                                         \\
		\multirow{-2}{*}{Method}                 & \multirow{-2}{*}{Selection criterion} & asym-40\%                    & sym-90\%                     & sym-80\%                  & sym-90\%                  \\ \hline
				                                 & best                                  & 91.16                        & 61.21                        & -                         & -                      \\
		\multirow{-2}{*}{PENCIL~\cite{Yi19}}     & last                                  & 91.01                        & 60.80                        & -                         & -                       \\ \hline
		                                         & best                                  & 93.4                         & 76.0                         & 60.2                      & 31.5                      \\
		\multirow{-2}{*}{DivideMix~\cite{LiJ20}} & last                                  & 92.1                         & 75.4                         & 59.6                      & 31.0                      \\ \hline
		                                         & best                                  & \textbf{94.6}                & 91.9                         & 66.4                      & 41.2                      \\
		\multirow{-2}{*}{AugDesc~\cite{Nishi21}} & last                                  & 94.3                         & 91.8                         & 66.1                      & 40.9                      \\ \hline
		                                         & best                                  & 93.95 $\pm$ 0.23             & \textbf{93.24 $\pm$ 0.47}    & 66.99 $\pm$ 0.48          & 54.48 $\pm$ 1.25          \\
		\multirow{-2}{*}{F\&F (Ours)}            & last                                  & 93.58 $\pm$ 0.17             & \textbf{93.24 $\pm$ 0.51}    & \textbf{67.08 $\pm$ 0.51} & \textbf{54.54 $\pm$ 1.36} \\ \hline
	\end{tabular}
	\caption{Comparison between F\&F and previous state-of-the-art methods in terms of accuracy (\%) on noisy CIFAR datasets from~\cite{Tanaka18}. 
	Results of 
	PENCIL, 
	DivideMix and AugDesc are copied from respectively papers. `asym' and `sym' are abbreviations for asymmetric and symmetric noise, respectively.}
	\label{table: legacy_noisy_cifar}
\end{table*}

\subsection{Results on Clothing1M}

We start reporting results from the real-world Clothing1M dataset, whose test set is \emph{imbalanced}. Different selection criteria were used to select the model: the one at the last training epoch, and the one with the best accuracy or the best mAP on the validation set. Note that no validation data is required for the `last' criterion. Results are in Table~\ref{table: clothing1m}.

It is clear that F\&F is the consistent winner in terms of mAP, the evaluation metric which we advocate because it does not sacrifice under-represented categories. But, when accuracy is used, F\&F is also a strong method. Even an incomplete F\&F (stopping at stage 1) consistently outperforms DivideMix. These results show that our warmup and self-supervised learning improvements in stage 1, and instance filtering and label correction in stage 2 are all effective.

Another interesting observation is that the `best mAP' criterion acts betters than the `best accuracy' one, which shows that accuracy is not an appropriate metric in imbalanced tasks from a different perspective.

\subsection{Results on nCIFAR}

On the proposed nCIFAR datasets, as shown in Table~\ref{table:nCIFAR}, the accuracy is used because its test set is balanced. We argue that a validation set is difficult to obtain in real world settings, and advocate using the last epoch's model. Table~\ref{table:nCIFAR} shows five time average of the accuracy of the last training epoch.

From Table~\ref{table:nCIFAR}, the stage 1 of F\&F consistently outperforms DivideMix, and the complete F\&F (stage 2) consistently beats its own stage 1.

Another interesting observation is that the complete F\&F has smallest variance in general, while F\&F stage 1 has smaller variance than DivideMix in most cases---meaning F\&F is more robust than DivideMix.

One of the reasons that methods which learn from noisy labels require a validation set is that they are sometimes not robust, hence a validation set is needed to pick the optimal model from the many training epochs. A robust method, like F\&F, suggests that we better \emph{fend off our dependency on the validation set}, and directly use the last epoch's model.

\subsection{Results on noisy CIFAR from~\cite{Tanaka18}}

We have discussed the drawbacks of the noisy CIFAR dataset previously used in the community, and have advocated the adoption of our nCIFAR datasets. However, to provide a thorough comparison to previous methods, we also conducted experiments on them, with results presented in Table~\ref{table: legacy_noisy_cifar}. We only experimented with the most difficult ones: 40\% asymmetric and 90\% symmetric noise on noisy CIFAR10, plus 80\% and 90\% symmetric noise on noisy CIFAR100 from~\cite{Tanaka18}.

First, F\&F is still very robust. In fact, in three out of four cases, F\&F's `last' outperforms `best'. Second, except on noisy CIFAR10 with 40\% asymmetric noise, F\&F is the clear winner. Third, in the two CIFAR100 symmetric noise cases, the winning margin of F\&F is significant, even more than 10 percentage points with 90\% symmetric noise.

For the last observation, our conjecture is that two factors lead to this large gap: self-supervised learning (especially our diverse initialization strategy), and label correction.

\section{Conclusions and Future Work}

In this paper, we first advocated two changes to the area of learning from examples with noisy labels: the dataset and the evaluation metric. We analyzed the drawbacks of the current datasets and metric, then proposed replacements: our nCIFAR datasets and the mAP metric (at least for imbalanced test sets). Then, we proposed an F\&F framework, which highlights four `friends' with our improved techniques: diverse self-supervised learning, freezed warmup, label correction and instance filtering. F\&F has not only outperformed existing methods, but also has higher robustness---its last epoch model's prediction has smaller variances.

The filed of learning from noisy labels shares similarity with learning from webly supervised data. In the future, we will migrate F\&F to such datasets as WebVision~\cite{Li17b}. Furthermore, learning from noisy labels in a multi-label setting is not only more challenging but also has wider utility in applications. We will investigate into this area by relaxing constraints on the soft labels in F\&F.


{\small
	\bibliographystyle{ieee_fullname}
	\bibliography{FaF_preprint}
}

\clearpage

\onecolumn
\appendix

\newcommand{\sharpen}[1]{\exp{\left( C_{#1} / \tau_{c} \right)}}
\newcommand{\deri}[2]{\frac{\mathrm{d} {#1}}{\mathrm{d} {#2}}}
\newcommand{\sharpenderi}[1]{{-\frac{C_{#1}}{\tau_c^2} \sharpen{#1}}}

\section{Proof of the monotonicity of the closed noise rate with respect to $\tau_c$}

As is mentioned in the paper, one entry $C_{ij}$ in the closed-world similarity matrix is calculated using the cosine similarity between the representation vectors of the $i$-th and $j$-th categories for $1 \le i,j \le n$. The cosine similarity between one vector and itself equals to 1, \ie, $C_{ii}= 1$ for any $1 \le i \le n$. Different categories always have different representation vectors, and hence $-1<C_{ij}<1$ if $i \neq j$. Thus, we can safely assume that $C_{ii} > C_{ti}$ holds for all $t \neq i$.

Now consider the set of closed-world examples generated in the proposed nCIFAR dataset with hyperparameter $\tau_c$, which includes both instances with clean and closed-world noisy labels. We denote the percentage of closed-world noisy labels as $r$ (\ie, the number of closed noise divided by the number of all closed-world examples). According to the approach that generates examples in nCIFAR, we have
\begin{equation}
	r = 1 - \frac{1}{n}{\sum_{i=1}^{n}{C_{ii}'}} \,,
\end{equation}
where 
\begin{equation}
   C_{ii}' = \frac{\sharpen{ii}}{\sum_{t=1}^{n}{\sharpen{ti}}}
\end{equation}
for $i = 1, 2, \ldots, n$. Note that we want $r = \frac{y}{1-x}$ in nCIFAR.

Then,
\begin{align}
   \deri{r}{\tau_c}
   &= -\frac{1}{n} \sum_{i=1}^{n} \deri{C_{ii}'}{\tau_c} \\
   &= -\frac{1}{n} \sum_{i=1}^{n} \deri{}{\tau_{c}} \left( \frac{\sharpen{ii}}{\sum_{t=1}^{n}{\sharpen{ti}}} \right) \\
   &= -\frac{1}{n} \sum_{i=1}^{n} \frac{\Big(\sum\limits_{t=1}^{n}{\sharpen{ti}}\Big) \cdot \left(\sharpenderi{ii}\right) - \left( \sum\limits_{t=1}^{n} \sharpenderi{ti}\right) \cdot \sharpen{ii}} {\left(\sum\limits_{t=1}^{n} \sharpen{ti}\right)^2} \\
   &= \frac{1}{n} \sum_{i=1}^{n} \frac{\sharpen{ii} \sum\limits_{t=1}^{n} (C_{ii} - C_{ti}) \sharpen{ti}} {\tau_c^2\left(\sum\limits_{t=1}^{n} \sharpen{ti}\right)^2} \\
   &> 0\,.
\end{align}
That is, the closed noise rate $r$ is monotonically increasing with respect to the temperature $\tau_{c}$. Therefore, it is reasonable to apply binary search to efficiently find the exact $\tau_{c}$ value that leads to $r = \frac{y}{1-x}$.

\section{Details of the loss function in our improved PENCIL}

It is known that PENCIL suffers from a seemingly unreasonable hyperparameter value, which is the learning rate used to update the soft labels $\vec{y}^d$. In F\&F, the key technique to solve this issue is: to update soft labels with a modified loss function, and to update soft labels and network parameters separately.

As is shown in the paper, the loss related to one sample $\vec{x}$ is defined as
\begin{equation}
   \mathrm{CE}(f(\vec{x}),\vec{y}^d)= - \sum_{i=1}^c f_i(\vec{x}) \log y^d_i \,.
\end{equation}

Both network parameters and soft labels are updated using SGD. Let us assume that the size of a mini-batch is denoted by $B$. For the $j$-th example in this mini-batch, the example, the network's prediction and its soft label are denoted by $\vec{x}_j$, $f(\vec{x}_{j})$ and $\vec{y}_{j}^d$, respectively. Note that the $i$-th dimension of the vector $\vec{y}_{j}^d$ is denoted as $y_{j,i}^d$.

We follow the conventional approach, and use the average gradient from a mini-batch to update network parameters, because the same set of network parameters are shared among all the samples. Hence, the loss term responsible for updating network parameters is 
\begin{equation}
	\mathcal{L}_1 = - \frac{1}{B} \sum_{j=1}^B \sum_{i=1}^c f_{i}(\vec{x}_j) \log y^d_{j,i} \,.
\end{equation}

When it turns to update the soft labels, the situation is completely different: one sample $\vec{x}_j$ has its own soft label $\vec{y}^d_j$; and, $\vec{y}^d_j$ is only corresponding to $\vec{x}_j$ and it is \emph{not} related to other $\vec{x}_t$ or $\vec{y}^d_t$ when $t \neq j$. Hence, the gradient must \emph{not} be averaged within a mini-batch. 

In PENCIL, the learning rate to update soft labels has to be enlarged to offset the effect of the batch size, namely the coefficient $\frac{1}{B}$. In other words, this hyperparameter is in fact \emph{dependent on the mini-batch size in PENCIL}. This explains why this hyperparameter has unusual values like 3000 or 10000 in PENCIL.

We use the following loss term to update soft labels:
\begin{equation}
	\mathcal{L}_2 = - \sum_{j=1}^B \sum_{i=1}^c f_{i}(\vec{x}_j) \log y^d_{j,i} \,.
\end{equation}
By removing the $\frac{1}{B}$ term in the loss function, we can use \emph{the same learning rate} to update soft labels in all our F\&F experiments, \ie, now it becomes very robust.

Technically, we detach the gradient to network parameters when updating soft labels, and vice versa.

\end{document}